\documentclass{article}
\usepackage[english]{babel}
\usepackage[letterpaper,top=2cm,bottom=2cm,left=3cm,right=3cm,marginparwidth=1.75cm]{geometry}
\usepackage{amsmath}
\usepackage{graphicx}
\usepackage[colorlinks=true, allcolors=blue]{hyperref}
\usepackage{setspace} % 添加此行以设置行间距
\usepackage{setspace} % 设置行间距
\usepackage{booktabs} % 提供 \toprule, \midrule, \bottomrule
\usepackage{array}    % 增强表格功能
\usepackage{tabularx} % 扩展表格功能
\usepackage{float}    % 提供 [H] 位置选项
\usepackage{caption}  % 增强标题格式
\usepackage{pifont}   % 提供 \checkmark 和 \ding 符号
\usepackage[table]{xcolor} % 提供颜色支持，特别是 \colorbox
\usepackage{dingbat}  % 提供额外的符号支持
\usepackage{amsmath}
\usepackage{amssymb}
\usepackage{indentfirst}
\title{RT-DETR++ for UAV Object Detection}
\author{
    Shufang Yuan \\
    School of Computer Science \& Technology, Huazhong University \\ 
    of Science and Technology, Wuhan, Hubei, China \\
    E-mail: \texttt{d202582018@hust.edu.cn}
}
\date{} % 可选：移除日期
\usepackage[numbers]{natbib}
\begin{document}
\maketitle

\begin{abstract}
Object detection in unmanned aerial vehicle (UAV) imagery presents significant challenges. Issues such as densely packed small objects, scale variations, and occlusion are commonplace. This paper introduces RT-DETR++, which enhances the encoder component of the RT-DETR model. Our improvements focus on two key aspects. First, we introduce a channel-gated attention-based upsampling/downsampling (AU/AD) mechanism. This dual-path system minimizes errors and preserves details during feature layer propagation. Second, we incorporate CSP-PAC during feature fusion. This technique employs parallel hollow convolutions to process local and contextual information within the same layer, facilitating the integration of multi-scale features. Evaluation demonstrates that our novel neck design achieves superior performance in detecting small and densely packed objects. The model maintains sufficient speed for real-time detection without increasing computational complexity. This study provides an effective approach for feature encoding design in real-time detection systems.

\end{abstract}

\section{Introduction}

Amidst the wave of technological revolution, drone technology, as an innovative aerospace tool, has demonstrated broad application prospects and immense development potential. Particularly in the field of aerospace and computer vision integration, drone aerial image target detection technology has become a hot research topic among numerous scholars. It holds significant academic value and practical significance in critical domains such as military operations, risk detection, and disaster prevention.

But images from UAVs\cite{zou2023object} have unique challenges compared to ground-level views. The flying height and camera angle can make target objects appear very small. Quick changes in flight path and altitude cause big variations in object scale and viewpoint. City textures, repeating patterns, and shadows can create background noise. Also, crowds and traffic are often dense, which means objects block each other. These issues make it hard for even the best object detectors to work well. They might not find small objects, miss targets in crowded areas, or struggle to define object boundaries correctly. Onboard computers on UAVs have limited power and processing ability. This puts strict limits on model size and how fast it can run. So, the main goal is to get the best detection quality for small, crowded targets with a limited computational budget.

In recent years, the development of target detection technology for drone aerial imagery has evolved from traditional detection methods to deep learning-based approaches. Traditional detection algorithms, such as Scale-Invariant Feature Transform, Histogram of Oriented Gradients, and deformable part models, typically rely on manually extracted features. These methods are not only time-consuming and labor-intensive but also sensitive to changes in image scale. When processing small objects captured from aerial perspectives, they struggle to efficiently extract detailed target information within a short timeframe, resulting in slow detection processes and insufficient accuracy. With the continuous advancement of deep learning, object detection algorithms based on convolutional neural networks (CNNs), such as, have achieved effective recognition of target objects at different scales through their complex network architectures that automatically learn multi-level feature representations. Demonstrating robust detection performance and generalization capabilities, these methods have gradually become a widely researched hotspot in academia.

 Object detection methods are usually split into two types: those based on convolutional neural networks (CNNs) and those based on Transformers. Faster R-CNN\cite{ren2015faster} was an important step forward. It used a region proposal network that made the detector more efficient and allowed for end-to-end training. But it still had trouble with objects of different sizes. Using a multi-scale image pyramid\cite{adelson1984pyramid, dalal2005histograms} is one way to handle scale changes, but it takes a lot of time because it has to process image features over and over. To address the challenge of detecting multi-scale objects in aerial images, Sun\cite{sun2024gd} proposed a novel feature fusion architecture called GDPAN. This architecture combines the traditional Path Aggregation Network (PANet)\cite{liu2018path} with a unique collection and distribution mechanism, effectively enhancing the model's ability to extract and fuse features for multi-scale objects. To effectively address the challenges of low resolution and recognition difficulties for small objects in dense scenes, Li Zihao\cite{li2023aerial} proposed the ACAM-YOLO adaptive collaborative attention mechanism detection algorithm. By employing a multi-region channel attention mechanism and adaptive attention weight fusion technology, this algorithm deeply extracts key contextual information, significantly enhancing the model's detection accuracy. However, when encountering severely occluded or feature-poor targets, this algorithm remains prone to false detections. However, it also increases the model's parameter count and computational complexity to some extent. To achieve lightweight model deployment on resource-constrained drone platforms, He Qitian\cite{zhang2023lightweight} integrated the FasterNet lightweight module and CBAM\_l attention mechanism into the YOLOv5 architecture. Their goal was to reduce parameter size and enhance detection accuracy through these improvements. However, while this strategy effectively decreased the model's parameter count, noticeable false negatives were still observed during the visualization and analysis of small-object detection. To address the issues of target omission and misclassification caused by changes in aerial photography perspectives during drone operations in complex environments, Li\cite{li2025pcaf} proposed a pyramid-based convergence and distribution fusion network. By progressively integrating multi-scale feature maps within the backbone network, this approach effectively promotes interaction between feature information at different scales, thereby reducing the probability of misclassification and omission. However, while the proposed network demonstrates strong performance under conventional conditions, its robustness requires further validation when dealing with image distortions caused by extreme weather conditions such as smog or atmospheric turbulence. Even with these improvements, CNN-based methods still find it difficult to detect small objects when they are dense and blocked.

In 2020, the Facebook AI Research team created the end-to-end object detection Transformer (DETR) model\cite{carion2020end}. It was the first end-to-end network to use Transformers for object detection. DETR had some clear benefits, but it also took a long time to train and converged slowly. To fix this, H-DETR\cite{jia2023detrs} developed a hybrid matching system that adjusts computing resources based on the task, keeping performance high while controlling cost. Group-DETR\cite{chen2023group} used a group-wise assignment method to speed up training. In 2024, Baidu released RT-DETR\cite{zhao2024detrs}, which performed very well in both accuracy and speed. It was quickly adapted for many different tasks, including small object recognition.

Although these end-to-end models look good, they can't actually handle the two most troublesome problems in drone detection. First, in the process of image zooming in, there will always be some important information that is inexplicably disappearing, and there will be more and more serious situations. Those methods commonly used today tend to ignore each other, either losing some key details, or they can't accurately understand the meaning of the image.capture and understanding of the overall environment. Second, the RT-DETR model suffers from insufficient multi-scale contextual modeling capabilities. Its convolutions feature a single dilation rate, preventing the model from effectively capturing both local details and global context. This limitation becomes particularly pronounced in scenarios involving complex backgrounds and small targets.

To address these challenges, we retained the real-time, end-to-end detection model architecture while focusing on two aspects of the encoder: sampling quality and multi-scale contextual integration. Our main contributions include:

(1) We developed an attention-guided up/down-sampling (AU/AD) module. This module employs a parallel dual-branch and channel-gating mechanism to reduce information loss and channel redundancy in drone images. This enhances feature transfer between layers.

(2) We also introduce a Cross-Stage Partially Parallel Hollow Convolution (CSP-PAC) fusion block. It leverages parallel hollow convolutions to capture multi-scale contextual information within the same layer. This design balances local details and global context, improving detection of small targets and occluded objects.

(3) Our evaluation on public drone datasets demonstrates that our approach achieves improved accuracy in both overall object and small object detection. Simultaneously, it enables real-time detection while maintaining sufficient efficiency. This indicates our design is both effective and practical for typical drone challenges.

\section{Methods}

This paper introduces RT-DETR++, an enhanced model based on RT-DETR. It retains the backbone network and Transformer decoder while optimizing components such as feature extraction and fusion in the encoder. By employing multi-head attention and multi-scale feature enhancement, the model improves detection performance. It is designed to address challenges in drone aerial images, including blurred object edges, difficulties in detecting small targets, and interference from complex backgrounds.

\subsection{Overall Framework}

\begin{figure}[t]
  \centering
  \includegraphics[width=\columnwidth]{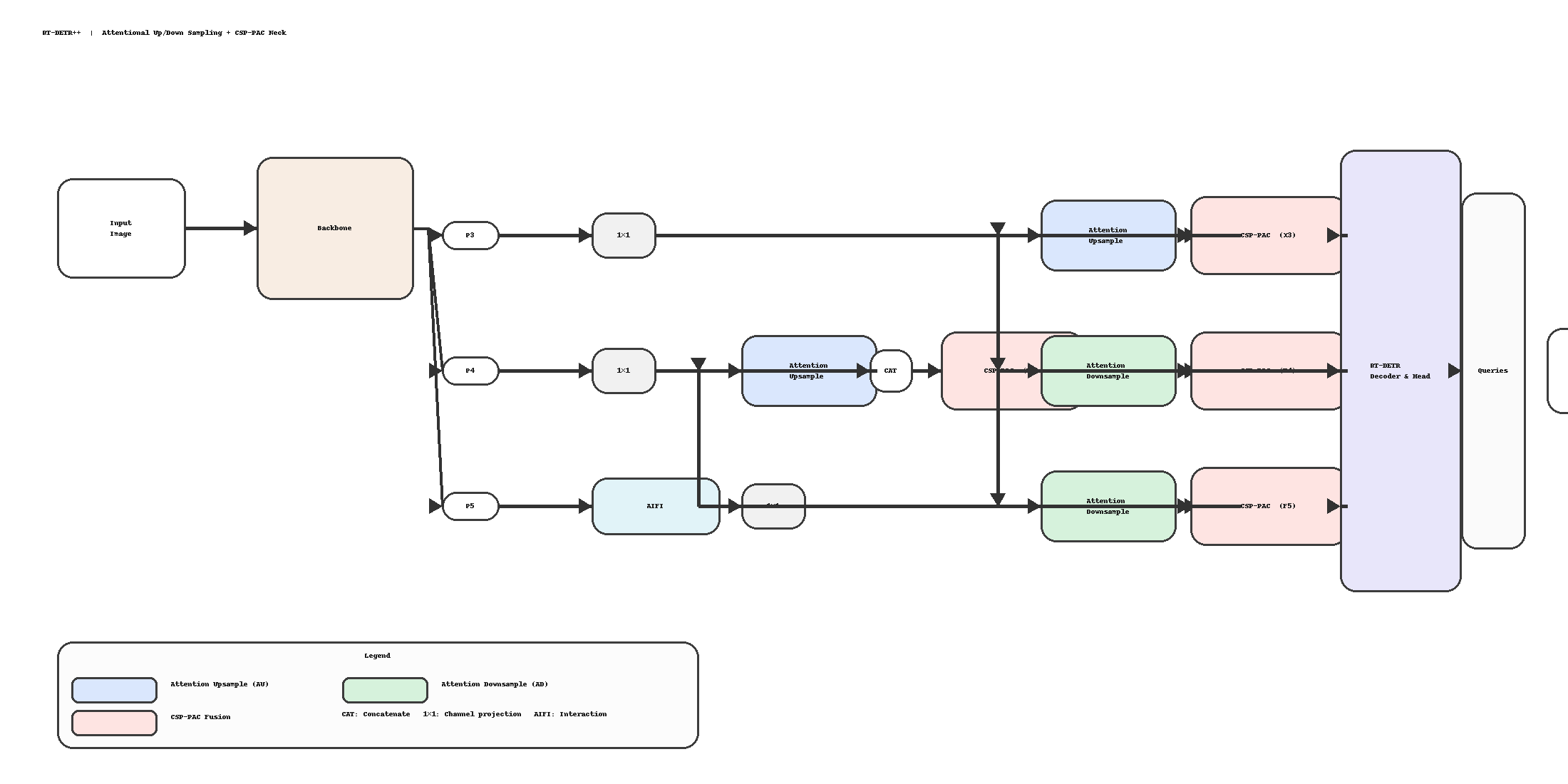}
  \caption{
    \textbf{The overall framework of RT-DETR++ model.} 
  }
  \label{FIG:1}
\end{figure}

Figure~\ref{FIG:1} shows the overall architecture. The RT-DETR++ model has three main parts: a Backbone, a Hybrid Encoder, and a Transformer Decoder. The backbone uses a ResNet-based structure to extract multi-scale features P3, P4, and P5. This paper adds several new modules to the feature pyramid network.

(1)An Attention Upsampling Module is used for top-down feature fusion. It has a dual-branch structure and a channel attention mechanism for content-aware feature improvement.

(2)An Attention Downsampling Module helps with bottom-up feature propagation. It uses a dual-path strategy to keep fine-grained details.

(3)A Parallel Atrous Convolution CSP (CSP-PAC) Module is placed in the feature fusion network. It uses parallel convolutions with different dilation rates to make the multi-scale feature representation stronger.

These modules work together to create a more efficient feature pyramid network. This greatly improves detection accuracy for multi-scale targets in UAV images while keeping the model fast enough for real-time.

\subsection{Attention Upsampling Module (AU) }

\begin{figure}[t]
  \centering
  \includegraphics[width=\columnwidth]{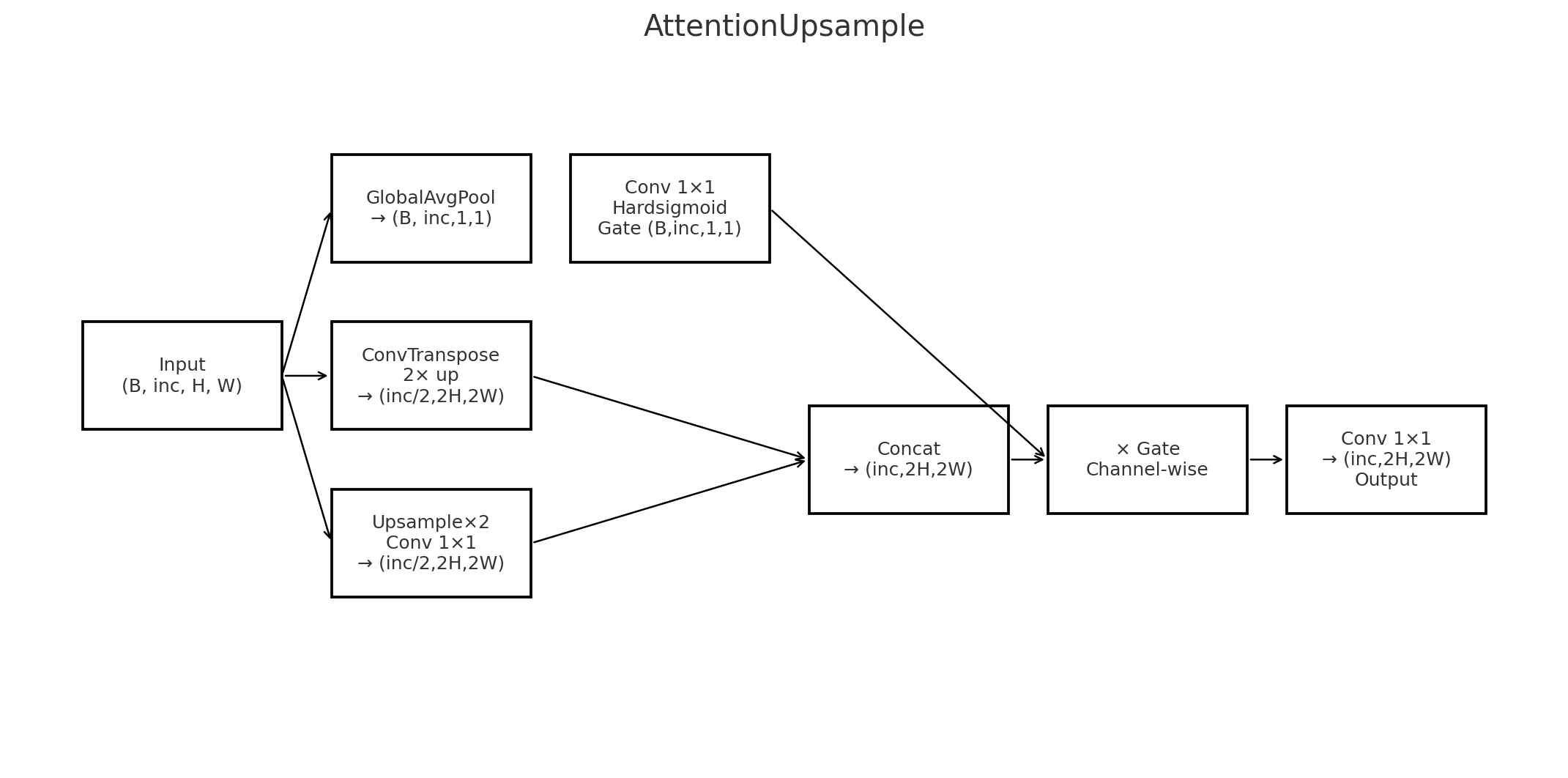}
  \caption{
    \textbf{The framework of the AU model.} 
  }
  \label{FIG:2}
\end{figure}

Traditional upsampling methods, such as nearest-neighbor interpolation, lack content-aware capabilities, resulting in feature maps that lack semantic consistency. To address this issue, we propose the AttentionUpsample module (Figure~\ref{FIG:2}). This module employs a dual-branch architecture and channel attention mechanism for feature upsampling, thereby achieving content-aware capabilities.

Given an input feature map $X \in \mathbb{R}^{C\times H\times W}$, the forward process of the module is defined as:

\begin{align*}
G &= \sigma\left(\mathrm{Conv}\left(\mathrm{AvgPool}(X)\right)\right) \in \mathbb{R}^{C\times 1\times 1} \\
U_1 &= \mathrm{ConvTranspose}(X, C/2, \mathrm{stride}=2) \\
U_2 &= \mathrm{Conv}\left(\mathrm{Upsample}(X, \mathrm{scale\_factor}=2), C/2\right) \\
X_{\mathrm{up}} &= \mathrm{Concat}(U_1, U_2) \odot G \\
Y &= \mathrm{Conv}(X_{\mathrm{up}}, C)
\end{align*}

Here, $G$ denotes the channel attention weights, $\sigma$ is the Sigmoid activation function, and $\odot$ represents element-wise multiplication.

This design combines the learning capabilities of transposed convolutions while maintaining the efficiency of simple interpolation. The channel attention mechanism adjusts channel weights based on feature content, thereby amplifying important features and suppressing redundant information. As demonstrated in the experimental section, this module enhances feature fusion quality with minimal additional computational overhead.

\subsection{Attention Downsampling Module (AD) }

\begin{figure}[t]
  \centering
  \includegraphics[width=\columnwidth]{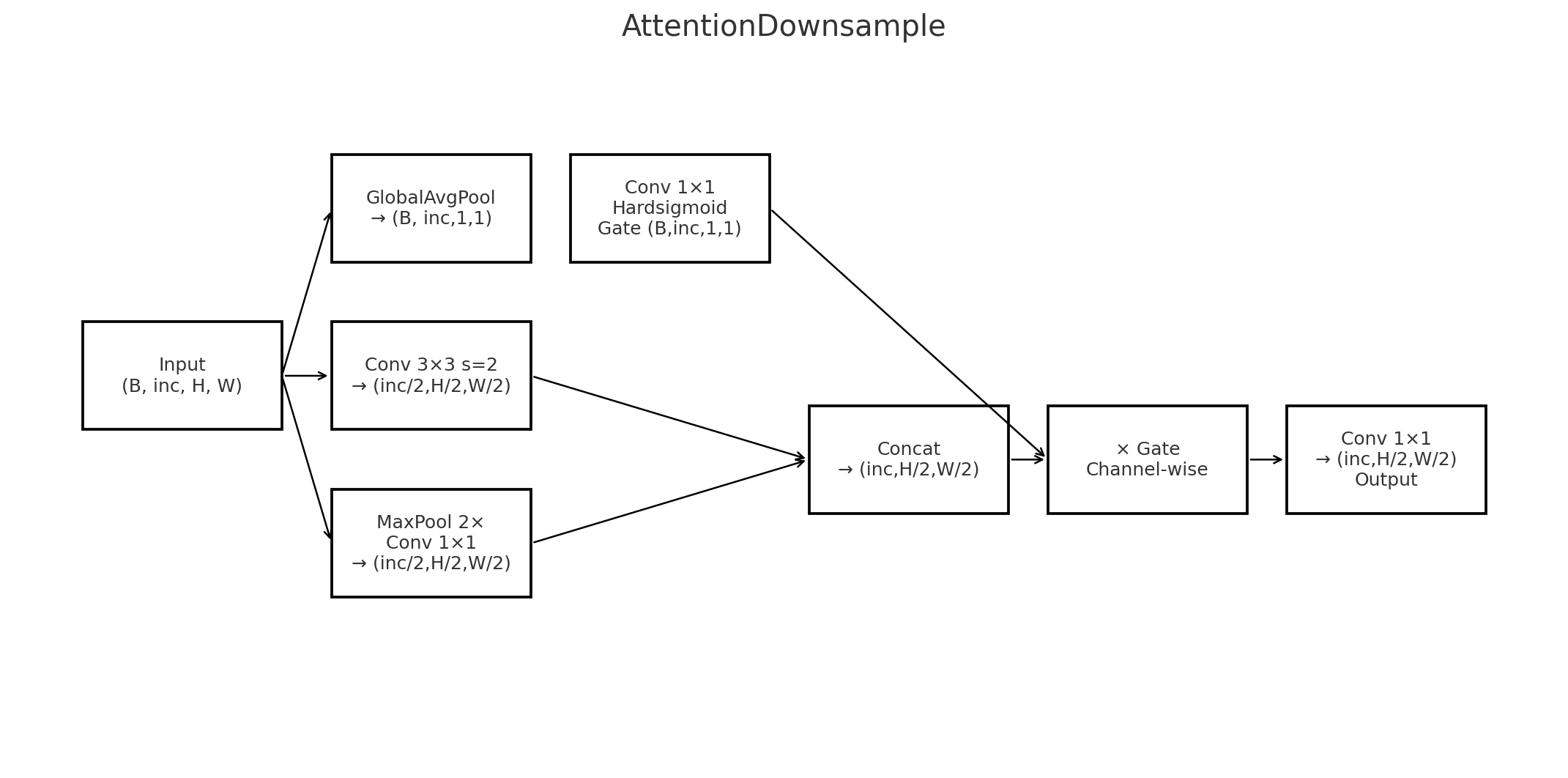}
  \caption{
    \textbf{The framework of the AD model.} 
  }
  \label{FIG:3}
\end{figure}

Downsampling is an important step in object detection, but common methods like max-pooling can cause a loss of fine-grained information. Our Attention Downsampling module is designed to keep important details in a content-aware way (Figure~\ref{FIG:3}).

Given an input feature map $X \in \mathbb{R}^{C\times H\times W}$, the forward process of the module is defined as:

\begin{align*}
G &= \sigma\left(\mathrm{Conv}\left(\mathrm{AvgPool}(X)\right)\right) \in \mathbb{R}^{C\times 1\times 1} \\
D_1 &= \mathrm{Conv}(X, C/2, \mathrm{kernel}=3, \mathrm{stride}=2) \\
D_2 &= \mathrm{Conv}\left(\mathrm{MaxPool}(X, \mathrm{kernel}=2, \mathrm{stride}=2), C/2\right) \\
X_{\mathrm{down}} &= \mathrm{Concat}(D_1, D_2) \odot G \\
Y &= \mathrm{Conv}(X_{\mathrm{down}}, C)
\end{align*}

The dual-path method preserves structural details while learning representative features. As demonstrated in the experimental section, this module effectively reduces information loss during downsampling while maintaining computational efficiency.

\subsection{Cross Stage Partial with Parallel Atrous Convolution (CSP-PAC) Module}

\begin{figure}[t]
  \centering
  \includegraphics[width=\columnwidth]{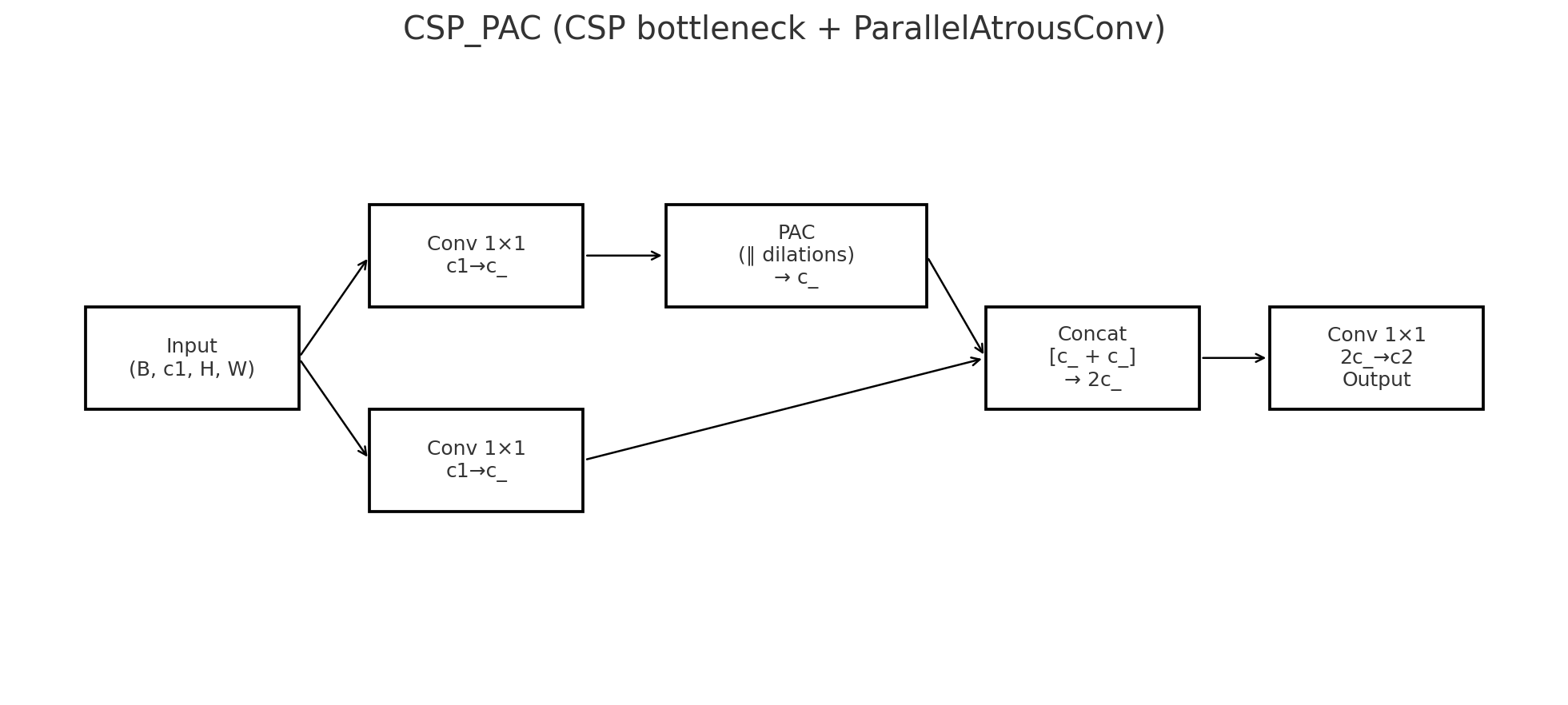}
  \caption{
    \textbf{The framework of the CSP-PAC model.} 
  }
  \label{FIG:4}
\end{figure}

The original Cross Stage Partial (CSP) bottleneck is good for feature extraction, but it doesn't explicitly model multi-scale context. We propose the CSP-PAC module (Figure~\ref{FIG:4}), which adds Parallel Atrous Convolution (PAC) to the CSP structure to improve multi-scale perception.

The PAC module is structured as:

\begin{align*}
\mathrm{PAC}(X) &= \mathrm{Conv}_{1\times 1}\left(\mathrm{Concat}\left(\mathrm{Conv}_{d=1}(X), \mathrm{Conv}_{d=2}(X), \mathrm{Conv}_{d=3}(X)\right)\right) \\
\end{align*}

Here, $\text{Conv}_{d=r}$ denotes a 3×3 convolution with dilation rate $r$.

The PAC module is embedded into the CSP bottleneck, forming the complete CSP-PAC design:

\begin{align*}
F_{\mathrm{main}} &= \mathrm{PAC}\left(\mathrm{Conv}_{1\times 1}(X)\right) \\
F_{\mathrm{shortcut}} &= \mathrm{Conv}_{1\times 1}(X) \\
Y &= \mathrm{Conv}_{1\times 1}\left(\mathrm{Concat}(F_{\mathrm{main}}, F_{\mathrm{shortcut}})\right)
\end{align*}

The PAC module is embedded into the CSP bottleneck. This design captures features at multiple receptive fields in a single module. It effectively handles the scale variation challenges in object detection. The PAC module is more computationally efficient and easier to optimize than deformable convolutions.

The CSP-PAC module replaces standard convolution operations and is employed in the feature fusion stage of the RT-DETR++ feature pyramid network. As demonstrated in the experimental section, this module significantly enhances the detection performance of multi-scale objects in crowded scenes and small-object scenarios.

\section{Experiment and result analysis}

\subsection{experimental environment}

The experiments were conducted on a system running Ubuntu 22.04, equipped with an NVIDIA GeForce RTX 4090D GPU, CUDA 11.3, and the PyTorch 1.11.0 framework, utilizing Python 3.8. All experiments employed identical software and hardware environments to ensure consistency. During model training, we followed the official best practices for the RT-DETR model to optimize training parameters, which helped improve training efficiency and model generalization capabilities.

\subsection{Dataset Introduction}

This study employs the VisDrone2019 dataset for training and testing, evaluating results on the VisDrone2019-DET-Test subset. VisDrone2019 is a large-scale dataset specifically created for drone vision research, comprising 10,209 images divided into a training set (6,471 images), validation set (548 images), and test set (3,190 images) . The images cover diverse complex scenes and ten object categories: pedestrians, people, bicycles, cars, vans, trucks, tricycles, tricycles with a canopy, buses, and motorcycles. The dataset includes over 2.6 million manually annotated object bounding boxes.

\subsection{Comparative experiment}

To understand the advancements of the RT-DETR++ model, we conducted a series of comparative experiments on the VisDrone DET test dataset. We also compared this method with other mainstream lightweight detectors to validate its performance and speed.

\begin{table}[H]
\footnotesize
\caption{%
% \parbox[t]{\textwidth}{%
\textbf{Comparison of AP (\%) and latency on the VisDrone validation set.} 
% }%
}
\label{tbl1}
\centering
\begin{tabular*}{\textwidth}{@{\extracolsep{\fill}}c|cccccc|cl}
\toprule
Model  & $AP$ & $AP_{50}$ & $AP_{75}$ & $AP_s$ & $AP_m$ & $AP_l$ &Latency (ms) \\
\midrule
YOLOv8-N\cite{sohan2024review}& 19.1 & 33.0 & 18.9 & 10.6 & 28.9 & 38.3 &  4.3  \\
YOLOv7-Tiny\cite{wang2023yolov7}& 19.4 & 35.1 & 18.5 & 10.5 & 29.1 & 41.0 &  3.8  \\
RemDet-Tiny\cite{li2025remdet}& 21.8 & 37.1 & 21.9 & 12.7 & 33.0 & \textbf{44.5} & 3.4 \\
RT-DETR-R18& 21.9 & 38.1 & 22.4 & 12.5 & 33.2 & 39.9 & 4.1 \\
\textbf{\textit{RT-DETR++(Ours)}}& \textbf{24.1} & \textbf{40.1} & \textbf{23.6} & \textbf{14.2} & \textbf{34.9} & 41.2 & 5.5 \\
\bottomrule
\end{tabular*}
\end{table}

As shown in Table \ref{tbl1}, RT-DETR++ performs better overall in UAV scenarios. Specifically, the baseline RT-DETR-R18 achieves an overall accuracy of 21.9\% AP, while RT-DETR++ improves it to 24.1\% AP,  representing an improvement of +2.2 AP.Regarding localization metrics, $AP_{50}$ and $AP_{75}$ improved from 38.1 and 22.4 to 40.1 and 23.6, respectively, indicating enhanced recall capability and higher localization accuracy at high IoU thresholds.Multi-scale performance observations reveal significant improvements in small object detection: $AP_s$ increases from 12.5 to 14.2, achieving a gain of +1.7 AP; accuracy for medium and large objects improved by 1.7 and 1.3 percentage points, respectively. This suggests that the proposed attention-aware sampling and parallel hollow convolution mechanism effectively mitigates the challenges of dense, weakly featured small object distributions in drone scenarios. Furthermore, despite RT-DETR++'s inference latency increasing by approximately 1.4 milliseconds to 5.5 milliseconds compared to the baseline, it remains within real-time capabilities.

\subsection{Ablation experiment}

We conducted incremental ablation experiments based on the RT-DETR-R18 model to validate the contribution of each module, with results summarized in Table \ref{tbl2}. First, introducing AttentionUpsample boosted the overall AP to 22.4 (+0.5), while slightly improving  $AP_s$, indicating that parallel upscaling and channel gating mechanisms effectively mitigate interpolation artifacts and enhance high-resolution features.AttentionDownsample further elevated AP to 22.7 and small object accuracy to 13.2, demonstrating that the parallel downsampling mechanism better preserves edge and salient region features in the bottom-up path, thereby improving detection performance in crowded and occluded scenarios. After integrating the CSP-PAC module, overall performance reached 23.6\% AP,with small object detection improving by 1.4 percentage points. Both $AP_{50}$ and $AP_{75}$increased by over 1 percentage point, demonstrating that parallel hollow convolutions enhance feature representation by introducing multi-scale context within the same layer. Ultimately, when all three modules work synergistically, RT-DETR++ achieves optimal performance: overall AP reaches 24.1, small object accuracy improves to 14.2, and detection rates for medium/large objects reach 34.9 and 41.2, respectively. In terms of computational complexity, the full model adds approximately 3.6 GFLOPs of floating-point operations and 1.5 million parameters compared to the baseline, while maintaining real-time inference latency.

\begin{table}[t]
\footnotesize
\setlength{\tabcolsep}{3.5pt} % ← 收紧列间距（默认 6pt）
\caption{%
\textbf{Comparison of AP (\%) of \emph{RT-DETR++} with different modules on the VisDrone validation set
(Baseline: RT-DETR-R18).} “\(\uparrow\)” indicates improvement over the baseline.
FLOPs are reported in GFLOPs; Params in millions.%
}
\label{tbl2}
\centering
\begin{tabular}{@{}ccc|cccccc|cc@{}} % ← 改为 tabular，并去掉左右边距
\toprule
AU & AD & CSP-PAC &
$AP$ & $AP_{50}$ & $AP_{75}$ & $AP_s$ & $AP_m$ & $AP_l$ &
FLOPs & Params \\
\midrule
\ding{53}  & \ding{53}  & \ding{53}  & 21.9 & 38.1 & 22.4 & 12.5 & 33.2  & 39.9 & 57.0 & 19.8 \\
\checkmark & \ding{53}  & \ding{53}  & 22.4 {\colorbox{cyan!15}{\(\uparrow\,0.5\)}}
                                       & 38.2 {\colorbox{cyan!15}{\(\uparrow\,0.1\)}}
                                       & 22.7 {\colorbox{cyan!15}{\(\uparrow\,0.3\)}}
                                       & 12.8 {\colorbox{cyan!15}{\(\uparrow\,0.3\)}}
                                       & 33.6 {\colorbox{cyan!15}{\(\uparrow\,0.4\)}}
                                       & 40.4 {\colorbox{cyan!15}{\(\uparrow\,0.5\)}} & 58.2 & 20.2 \\
\checkmark & \checkmark & \ding{53}  & 22.7 {\colorbox{cyan!15}{\(\uparrow\,0.8\)}}
                                       & 38.9 {\colorbox{cyan!15}{\(\uparrow\,0.8\)}}
                                       & 23.0 {\colorbox{cyan!15}{\(\uparrow\,0.6\)}}
                                       & 13.2 {\colorbox{cyan!15}{\(\uparrow\,0.7\)}}
                                       & 33.9 {\colorbox{cyan!15}{\(\uparrow\,0.7\)}}
                                       & 40.7 {\colorbox{cyan!15}{\(\uparrow\,0.8\)}} & 58.8 & 20.6 \\
\checkmark & \ding{53}  & \checkmark & 23.6 {\colorbox{cyan!15}{\(\uparrow\,1.7\)}}
                                       & 39.6 {\colorbox{cyan!15}{\(\uparrow\,1.5\)}}
                                       & 23.2 {\colorbox{cyan!15}{\(\uparrow\,0.8\)}}
                                       & 13.9 {\colorbox{cyan!15}{\(\uparrow\,1.4\)}}
                                       & 34.4 {\colorbox{cyan!15}{\(\uparrow\,1.2\)}}
                                       & 41.0 {\colorbox{cyan!15}{\(\uparrow\,1.1\)}} & 59.2 & 20.9 \\
\checkmark & \checkmark & \checkmark & 24.1 {\colorbox{cyan!15}{\(\uparrow\,2.2\)}}
                                       & 40.1 {\colorbox{cyan!15}{\(\uparrow\,2.0\)}}
                                       & 23.6 {\colorbox{cyan!15}{\(\uparrow\,1.2\)}}
                                       & 14.2 {\colorbox{cyan!15}{\(\uparrow\,1.7\)}}
                                       & 34.9 {\colorbox{cyan!15}{\(\uparrow\,1.7\)}}
                                       & 41.2 {\colorbox{cyan!15}{\(\uparrow\,1.3\)}} & 60.6 & 21.3 \\
\bottomrule
\end{tabular}
\end{table}

\subsection{Experimental Summary}

In short, the comparative experiments show that RT-DETR++ is better than other methods in both overall accuracy and small object detection. It performs well while staying fast enough for real-time use. The ablation studies show how the different modules work together. CSP-PAC provides the biggest accuracy boost. AttentionUpsample and AttentionDownsample improve feature quality in the top-down and bottom-up paths. When combined, the model effectively uses multi-scale context and key features in UAV images. This leads to solid improvements in detecting small objects.

\section{Conclusion}

This paper addresses challenges in drone visual detection, such as dense small targets and complex occlusions, by proposing an enhanced RT-DETR model named RT-DETR++. The model employs a dual-branch parallel design introduced during upsampling and downsampling stages, combined with a channel gating mechanism, effectively mitigating information loss and artifact interference during cross-layer feature transmission. Furthermore, the CSP-PAC module integrates multi-scale dilated convolutions, enabling a single feature layer to capture both local details and richer contextual information. Experimental results on the VisDrone dataset demonstrate superior performance over baseline approaches, particularly in small object detection accuracy.

However, the current model still exhibits limitations under extreme conditions, such as poor performance when detecting objects smaller than 4 pixels or in heavily occluded scenes, and it incurs higher inference latency compared to baseline models. Future research could explore more lightweight contextual modeling mechanisms. Additionally, the generalization capability of this method requires validation through more diverse drone datasets.
In summary, RT-DETR++ enhances object detection performance in aerial scenarios while maintaining a concise detection paradigm, offering valuable insights for future research.

\bibliographystyle{unsrt}

\end{document}